\documentclass{article} 
\usepackage{iclr2026_conference,times}

\usepackage{amsmath,amsfonts,bm}

\def\eqref#1{equation~\ref{#1}}

\def\1{\bm{1}}

\DeclareMathAlphabet{\mathsfit}{\encodingdefault}{\sfdefault}{m}{sl}
\SetMathAlphabet{\mathsfit}{bold}{\encodingdefault}{\sfdefault}{bx}{n}

\newcommand{\resultrow}[8]{#1 & #2 & #4 & #5 & #6 & #7 & #8 & #3 \\}

\newcommand{\loss}{\mathcal{L}}

\usepackage{hyperref}
\usepackage{url}
\usepackage{graphicx} 
\usepackage{booktabs}
\usepackage{subcaption}
\usepackage{placeins}

\title{Revisiting the learning objectives of vision-language reward models}

\author{Simon Roy%
\thanks{Equal contribution. Corresponding author: simon-7.roy@etud.polymtl.ca}
~$^{1}$, Samuel Barbeau$^{*2}$,
Giovanni Beltrame$^{1}$, Christian Desrosiers$^{2}$, Nicolas Thome$^{3}$
    \vspace*{1mm} \\
    $^1$ Polytechnique Montréal, $^2$ École de Technologie Supérieure, $^3$ Sorbonne Université \vspace*{1mm} \\
 }

\newcommand{\ppm}[1]{\,\scriptsize{$\pm$#1}}

\iclrfinalcopy 
\begin{document}

\maketitle

\begin{abstract}

Learning generalizable reward functions is a core challenge in embodied intelligence. Recent work leverages contrastive vision language models (VLMs) to obtain dense, domain-agnostic rewards without human supervision. These methods adapt VLMs into reward models through increasingly complex learning objectives, yet meaningful comparison remains difficult due to differences in training data, architectures, and evaluation settings. In this work, we isolate the impact of the learning objective by evaluating recent VLM-based reward models under a unified framework with identical backbones, finetuning data, and evaluation environments. Using Meta-World tasks, we assess modeling accuracy by measuring consistency with ground truth reward and correlation with expert progress. Remarkably, we show that a simple triplet loss outperforms state-of-the-art methods, suggesting that much of the improvements in recent approaches could be attributed to differences in data and architectures.


\end{abstract}

\section{Introduction}

The increasing momentum of embodied intelligence has motivated the study of generalizable reward models that do not rely on hand-crafted human supervision. Recent work leverages large vision-language models (VLMs) as general-purpose reward functions that measure progress through alignment between visual observations and language goals. \cite{rocamonde2024visionlanguagemodelszeroshotreward} showed that CLIP \citep{radford2021learningtransferablevisualmodels} can be used as a zero-shot reward model for downstream policy learning, though it struggles with domain shift related to action specification and environmental dynamics. To mitigate this effect, many methods have turned to large-scale video demonstration datasets \citep{damen2018scalingegocentricvisionepickitchens, grauman2022ego4dworld3000hours} to finetune VLMs, enabling a better understanding of task-relevant behaviors. Nevertheless, the question of extracting meaningful signals of task progress from video demonstrations remains a central challenge to vision-language reward modeling. As a result, increasingly complex learning objectives have been proposed to generate more accurate rewards \citep{sermanet2018timecontrastivenetworksselfsupervisedlearning, nair2022r3muniversalvisualrepresentation, ma2023vipuniversalvisualreward, ma2023livlanguageimagerepresentationsrewards, karamcheti2023languagedrivenrepresentationlearningrobotics}. However, most of these methods were pre-trained using different datasets and architectures, making it difficult to isolate the choice of learning objective from downstream performance. For instance, R3M \citep{nair2022r3muniversalvisualrepresentation} was pre-trained on Ego4D \citep{grauman2022ego4dworld3000hours} yet compares to methods pre-trained on ImageNet \citep{russakovsky2015imagenetlargescalevisual,parisi2022unsurprisingeffectivenesspretrainedvision}. LIV \citep{ma2023livlanguageimagerepresentationsrewards} was initialized with CLIP weights and pre-trained on EpicKitchen \citep{nair2022r3muniversalvisualrepresentation} yet compares to methods utilizing a frozen BERT encoder \cite{devlin2019bertpretrainingdeepbidirectional} and vision encoder pre-trained on Ego4D from scratch \citep{nair2022r3muniversalvisualrepresentation}.


In this work, we systematically compare objectives under a unified framework, holding the pre-trained model backbone, finetuning data, and downstream evaluation environments constant. This setup allows us to decouple the impact of the learning objectives from any other confounding factors. We assess model performance via two distinct benchmarks, evaluating consistency with ground truth reward and alignment with expert progress. While initially introduced as a minimal baseline for comparison with recent methods, our results show that a very simple triplet loss \cite{Schroff_2015} can surpass current state-of-the-art learning objectives. More broadly, our findings indicate that simpler ranking-based learning objectives offer greater accuracy and robustness, suggesting that much of the apparent progress in recent methods could instead be due to differences in data and model architecture.

\section{Methodology} 
\label{learning_objectives}


State-of-the-art contrastive reward modeling approaches rely on combinations of learning objectives, either in their original form or adapted for alignment with language. For our experiments, we focus on a set of loss functions that can be used to reproduce many popular reward models. 

\noindent\textbf{TCN} \citep{sermanet2018timecontrastivenetworksselfsupervisedlearning} reduces the embedding distance of images which are closer in time and increases the distance of images which are temporally further apart. Given an image encoder $\pi_{img}$, a sequence of images $\{I_t\}_{t=0}^k$, their corresponding encodings $z_t=\pi_{img}(I_{t})$, batches of $[I_i, I_{j>i}, I_{k > j}]^{1:B}$ and a similarity function $\mathcal{S}$, TCN minimizes the following objective:
\begin{equation}
    \loss_\text{TCN} = - \frac{1}{B}\sum_{b \in B} \log \frac{e^{\mathcal{S}(z^b_i, z^b_j)}}{e^{\mathcal{S}(z^b_i, z^b_j)} + e^{\mathcal{S}(z^b_i, z^b_k)} + e^{\mathcal{S}(z^b_i, {z^{\neq b}_i})}} 
\label{eq:tcn}
\end{equation}

We adapt $\loss_\text{TCN}$ to language by replacing $I_{k > j}$ with the language annotation for the given sequence. Inserting the language embedding $v=\pi_\text{text}(l)$ results in the following objective 
\begin{equation}
    \loss_{\text{TCN}_{\text{text}}} = -\frac{1}{B}\sum_{b \in B} \log \frac{e^{\mathcal{S}(z^b_{j>i}, v^b)}}{e^{\mathcal{S}(z^b_{j>i}, v^b)} + e^{\mathcal{S}(z^b_{i}, v^b)} + e^{\mathcal{S}(z^{\neq b}_{j>i}, v^b)}}
\label{eq:lang}
\end{equation}

\noindent\textbf{VIP} \citep{ma2023vipuniversalvisualreward} uses a goal-conditioned approach that separates embeddings of temporally adjacent images while bringing together embeddings of images that are farther apart in time. Given batches of $[I_i, I_{j>i}, I_{j+1},I_{k\geq j+1}]^{1:B}$, VIP is defined as:
\begin{equation}
\loss_{\text{VIP}}=\frac{1-\gamma}{B} \sum_{b \in B} \left[-\mathcal{S}(z^b_i,z^b_k)\right] + \log \frac{1}{B} \sum_{b \in B} \mathrm{exp}\left[\mathcal{S}(z^b_j,z^b_k) +1 - \gamma \mathcal{S}(z^b_{j+1},z^b_k)\right]
\end{equation}

Just like TCN, we adapt $\loss_\text{VIP}$ to language by replacing $I_{k > h}$ with the embedded language annotation $v=\pi_\text{text}(l)$ for the given sequence.
\begin{equation}
\loss_{\text{VIP}_{\text{text}}}=\frac{1-\gamma}{B} \sum_{b \in B} \left[-\mathcal{S}(z^b_i,v^b)\right] + \log \frac{1}{B} \sum_{b \in B} \mathrm{exp}\left[\mathcal{S}(z^b_j,v^b) +1 - \gamma \mathcal{S}(z^b_{j+1},v^b)\right]
\end{equation}

\noindent\textbf{Combinations} of these losses reproduce popular reward modeling methods minus regularization terms. R3M is obtained by combining both TCN versions:
\begin{equation}
\loss_\text{R3M}=\loss_\text{TCN}+\loss_{\text{TCN}_{\text{text}}}
\end{equation}
Note that R3M adapts TCN to language by training an MLP to predict a similarity score from the concatenated vector [$z_0, z_i, v$], where $z_0$ is the first image in the sequence. This approach introduces a non-negligible amount of additional parameters, especially for large embeddings like those from SigLIP2. For fairness, we omit the initial embedding in our setup.

LIV is a combination of both VIP versions plus an InfoNCE objective \citep{oord2018representation}, the latter which we also include in our experiments:
\begin{equation}
\loss_\text{LIV}=\loss_\text{VIP}+\loss_{\text{VIP}_{\text{text}}}+\frac{1}{B} \sum_{b \in B} \left[-\log \frac{ e^{\mathcal{S}(z^b_k,v^b)}}{\frac{1}{B} \sum_{j \neq b}^B\left[e^{\mathcal{S}(z^j_k,v^b)}\right]} \right]
\end{equation}

\noindent\textbf{Triplet} \citep{Schroff_2015} applied to this context reduces the embedding distance between the language goal and the later images of the sequence, and pushes earlier images further away. We propose it as a simple baseline for task progress and show its remarkable effectiveness. Given batches of $[I_i, I_{j>i}, l]^{1:B}$, we use the language annotation as the anchor, the later image as the positive and the earlier image as the negative example. The triplet loss is defined as:
\begin{equation}
\loss_{\text{Triplet}} = \sum_{b \in B} \max \Big( 0, \mathcal{S}(v^b, x_i^b) - \mathcal{S}(v^b, x_j^b) + \alpha \Big)
\end{equation}
where $\alpha$ is the margin controlling how far apart the negative should be from the positive.

\section{Experiments}

\subsection{Finetuning}

\noindent\textbf{Learning objectives.} We finetune using different configurations of the losses described in \ref{learning_objectives}
\begin{center}
\begin{tabular}{@{}l l l l l l l@{}}
(1) $\loss_{\text{Triplet}}$ {\small(ours)} & 
(2) $\loss_{\text{TCN}_{\text{text}}}$ & 
(3) $\loss_\text{R3M}$ & 
(4) $\loss_{\text{VIP}_{\text{text}}}$ & 
(5) $\loss_{\text{VIP}_{\text{text}}}+\loss_\text{VIP}$ & 
(6) $\loss_\text{LIV}$
\end{tabular}
\end{center}

\noindent\textbf{Model backbone} To isolate the performance of the learning objective from model capacity and pre-training, we employ the base version of SigLIP2 \citep{tschannen2025siglip2multilingualvisionlanguage} as the backbone for all methods. We finetune using LoRA \citep{hu2021loralowrankadaptationlarge} to prevent overfitting.

\noindent\textbf{Data} Each model is finetuned on expert demonstrations from the Meta-World environments \citep{yu2021metaworldbenchmarkevaluationmultitask}. To evaluate out-of-domain performance, we remove 3 tasks from the training data (see \ref{evaluation}). For the remaining tasks, we collect 3 expert trajectories per task with randomized end-effector initializations, yielding 132 total. To ensure consistent dataset size, we cap the maximum number of samples at 50 thousand and generate them evenly across trajectories. Each timestep is recorded from 3 camera views to enable multi-view training. The timesteps within a sample remain fixed between epochs, but their associated views are randomly reassigned at each iteration, acting as a form of data augmentation to promote view-invariant representations and reduce overfitting.

\noindent\textbf{Implementation details} For a complete list of the implementation details, refer to \ref{hyperparameters}.

\subsection{Evaluation}
\label{evaluation}

We wish to learn generalizable reward models which can leverage pre-training knowledge to predict progress on unseen tasks. To this end, we exclude \texttt{button-press}, \texttt{drawer-open}, and \texttt{door-open} environments from the finetuning data and use them for subsequent evaluation metrics. \texttt{Button-press} is relatively simple, as success depends mainly on end-effector proximity to a target. In contrast, \texttt{drawer-open} and \texttt{door-open} are compositional, requiring navigation to the object and subsequent manipulation. Note that we evaluate only the standard \texttt{button-press} but remove all its variants to prevent leakage. 

To evaluate the performance of finetuned reward models on out-of-domain tasks, we introduce two benchmarks. \textbf{(1)} For measuring reward modeling robustness, we collect random and sub-optimal rollouts from each test environment to construct per-task datasets of 10,000 timestep pairs, each labeled by which timestep has the higher ground-truth reward. The model predicts the higher-reward timestep by comparing each element’s similarity to the language goal. Accuracy is defined as the percentage of pairs where this prediction matches the ground-truth ordering. We refer to this benchmark as \textit{consistency with ground-truth reward}. While informative about robustness, this metric does not necessarily indicate that a model can guide an agent towards a goal. \textbf{(2)} To evaluate this capacity, we collect 50 expert trajectories per task and predict rewards along each trajectory. Assuming expert timesteps correlate with increasing ground-truth reward, we compute the Value-Order Correlation (VOC) \cite{ma2024visionlanguagemodelsincontext} to measure how well the model’s predicted reward ordering aligns with goal-directed behavior. A model should perform well on both these benchmarks to yield transferable rewards. 

\section{Results}

We report all results on a per-view basis and multi-view. The multi-view is obtained by taking the average similarity scores across views of a timestep, and using it for prediction instead. This evaluates if the objectives were able to take advantage of the multi-view augmentations to learn view-agnostic representations.

\begin{table}[th]
\centering
\caption{Consistency with ground truth reward reported as prediction accuracy over random pairs.}
\label{tab:task_order}
\resizebox{0.8\linewidth}{!}{%
\begin{small}
\begin{tabular}{@{}l c c c c c c c@{}} 
\resultrow{}
{SigLIP2}
{\textbf{$\loss_{\text{Triplet}}$ {\footnotesize (ours)}}}
{\textbf{$\loss_{\text{TCN}_{\text{text}}}$}}
{\textbf{$\loss_\text{R3M}$}}
{\textbf{$\loss_{\text{VIP}_{\text{text}}}$}}
{\textbf{$\loss_{\text{VIP}_{\text{text}}}+\loss_\text{VIP}$}}{\textbf{$\loss_\text{LIV}$}}
\midrule
\multicolumn{8}{c}{\textbf{Button press}}\\
\midrule
\resultrow{View 1}{45.15}{\textbf{75.36}}{69.63}{74.42}{57.07}{57.09}{55.29}
\resultrow{View 2}{44.81}{\textbf{74.15}}{69.51}{71.38}{57.09}{56.18}{54.28}
\resultrow{View 3}{44.83}{\textbf{74.29}}{70.27}{72.07}{56.03}{53.69}{54.79}
\resultrow{Multi view}{40.50}{\textbf{76.44}}{72.24}{73.76}{57.83}{55.31}{55.51}
\midrule
\multicolumn{8}{c}{\textbf{Open drawer}}\\
\midrule
\resultrow{View 1}{42.91}{\textbf{69.01}}{64.52}{65.71}{65.58}{61.76}{50.56}
\resultrow{View 2}{41.55}{\textbf{67.82}}{61.83}{63.25}{60.45}{59.88}{50.77}
\resultrow{View 3}{43.60}{\textbf{66.33}}{61.73}{61.93}{58.49}{56.84}{49.51}
\resultrow{Multi view}{42.13}{\textbf{67.01}}{62.26}{62.24}{56.43}{55.25}{47.88}
\midrule
\multicolumn{8}{c}{\textbf{Open door}}\\
\midrule
\resultrow{View 1}{50.02}{62.78}{\textbf{64.80}}{61.04}{53.17}{49.42}{49.28}
\resultrow{View 2}{55.04}{64.27}{\textbf{65.46}}{62.33}{56.80}{54.22}{50.37}
\resultrow{View 3}{54.44}{\textbf{64.04}}{63.98}{63.64}{58.40}{55.04}{52.44}
\resultrow{Multi view}{57.27}{\textbf{65.02}}{64.96}{65.02}{60.12}{58.57}{54.21}
\midrule
\resultrow{Average}{46.85}{\textbf{68.88}}{65.93}{66.40}{58.12}{56.10}{52.07}
\bottomrule
\end{tabular}
\end{small}
}
\end{table}

\noindent\textbf{Consistency with ground truth reward} Table~\ref{tab:task_order} summarizes each model’s task consistency with ground-truth reward accuracy. Triplet loss achieves the highest overall accuracy across all held-out tasks, surpassing both TCN-based and VIP-based objectives. While $\loss_{\text{TCN}_{\text{text}}}$ and $\loss_\text{R3M}$ perform similarly, objectives incorporating VIP components exhibit inconsistent ranking ability, often close to random chance. The base SigLIP2 backbone performs slightly below 50\%, confirming that finetuning on expert demonstrations is necessary to encode temporal semantics.

\noindent\textbf{Alignment with expert progress} As shown in Table~\ref{tab:voc}, VOC scores reveal a similar trend. The triplet objective demonstrates strong correlations with expert progress, surpassing VIP based objectives on all tasks and rivaling TCN based objectives on \texttt{button-press} and \texttt{drawer-open}. In contrast, VIP based objectives exhibit high variance and frequent negative correlations even on the easier \texttt{button-press}, indicating reward inconsistencies along expert trajectories. All methods collapse on \texttt{door-open}, indicating a need for methods more adapted to multi-step tasks. Overall, these findings still support the conclusion that simpler ranking-based losses are capable of producing accurate and robust reward models. See \ref{reward_over_trajectories} for visualizations of predicted reward over expert trajectories for each model.

\begin{table}[th]
\centering
\caption{Alignment with expert progress reported as VOC scores (percentage).}
\label{tab:voc}
\resizebox{0.95\linewidth}{!}{%
\begin{small}
\def\arraystretch{0.95}
\begin{tabular}{@{}l c c c c c c c@{}} 
\resultrow{}
{SigLIP2}
{\textbf{$\loss_{\text{Triplet}}$ {\footnotesize (ours)}}}
{\textbf{$\loss_{\text{TCN}_{\text{text}}}$}}
{\textbf{$\loss_\text{R3M}$}}
{\textbf{$\loss_{\text{VIP}_{\text{text}}}$}}
{\textbf{$\loss_{\text{VIP}_{\text{text}}}+\loss_\text{VIP}$}}
{\textbf{$\loss_\text{LIV}$}}
\midrule
\multicolumn{8}{c}{\textbf{Button press}}\\
\midrule
\addlinespace[0.5ex]
\resultrow{View 1}{-72.80\ppm{10.87}}{\bf95.47\ppm{2.11}}{89.08\ppm{5.80}}{81.78\ppm{2.78}}{27.39\ppm{24.04}}{-21.03\ppm{13.11}}{16.67\ppm{17.89}}
\resultrow{View 2}{-19.50\ppm{17.28}}{90.03\ppm{3.28}}{90.92\ppm{3.43}}{\bf{90.95}\ppm{2.98}}{54.95\ppm{16.77}}{56.60\ppm{19.61}}{40.71\ppm{24.08}}
\resultrow{View 3}{-11.51\ppm{17.60}}{81.72\ppm{3.27}}{\bf82.55\ppm{5.20}}{79.10\ppm{4.04}}{77.94\ppm{9.55}}{63.89\ppm{9.89}}{32.12\ppm{16.22}}
\resultrow{Multi view}{-56.30\ppm{13.40}}{\bf95.07\ppm{2.29}}{94.53\ppm{2.41}}{86.87\ppm{2.81}}{68.21\ppm{15.57}}{56.72\ppm{14.08}}{42.9\ppm{15.35}}
\midrule
\multicolumn{8}{c}{\textbf{Open drawer}}\\
\midrule
\resultrow{View 1}{-56.76\ppm{29.20}}{87.67\ppm{3.90}} {82.17\ppm{5.21}} {\bf{87.70}\ppm{3.90}}{65.68\ppm{8.22}}{54.28\ppm{14.35}}{61.99\ppm{6.56}}
\resultrow{View 2}{-29.86\ppm{24.33}}{76.94\ppm{4.26}} {80.92\ppm{4.89}} {\bf83.71\ppm{4.75}}{69.87\ppm{17.17}}{53.92\ppm{31.42}}{66.99\ppm{10.04}}
\resultrow{View 3}{19.7\ppm{28.93}}{\bf86.76\ppm{4.12}} {75.88\ppm{12.34}} {80.88\ppm{7.01}}{67.54\ppm{13.45}}{64.17\ppm{12.27}}{55.47\ppm{16.78}}
\resultrow{Multi view}{-28.84\ppm{37.83}}{\bf90.17\ppm{2.26}} {86.14\ppm{2.80}} {87.22\ppm{3.25}}{77.59\ppm{9.53}}{75.17\ppm{9.33}}{80.40\ppm{5.39}}
\midrule
\multicolumn{8}{c}{\textbf{Open door}}\\
\midrule
\resultrow{View 1}{-14.8\ppm{20.33}}{\bf26.01\ppm{26.31}}{-42.27\ppm{15.58}}{2.21\ppm{22.80}}{-25.8\ppm{12.50}}{-0.04\ppm{17.66}}{17.51\ppm{17.43}}
\resultrow{View 2}{7.84\ppm{18.36}}{22.11\ppm{22.34}}{\bf42.66\ppm{23.50}}{7.7\ppm{17.50}}{-33.03\ppm{15.55}}{-15.03\ppm{15.53}}{-5.99\ppm{14.20}}
\resultrow{View 3}{29.52\ppm{17.17}}{-7.01\ppm{16.21}}{31.43\ppm{21.83}}{\bf33.7\ppm{23.65}}{-30.36\ppm{14.86}}{-27.62\ppm{13.04}}{-0.28\ppm{10.80}}
\resultrow{Multi view}{11.82\ppm{17.89}}{6.70\ppm{17.91}}{10.40\ppm{17.69}}{\bf14.33\ppm{20.68}}{-31.39\ppm{13.38}}{-18.94\ppm{11.91}}{-0.64\ppm{10.23}}
\midrule
\resultrow{\textbf{Average}}{-18.46\ppm{21.10}}{\bf62.64\ppm{9.02}}{60.36\ppm{10.06}}{61.35\ppm{9.68}}{32.38\ppm{14.22}}{28.51\ppm{15.18}}{33.99\ppm{13.75}}
\bottomrule
\end{tabular}
\end{small}
}
\end{table}

\section{Conclusion \& Future Work}


In this work, we revisit the learning objectives of recent vision-language reward models and evaluate them under a unified framework controlling for architecture, data, and training conditions. We find that a simple triplet loss can outperform more complex objectives in robustness to sub-optimal trajectories, and that ranking-based objectives correlate more strongly with ground-truth rewards. These results underscore the need for standardized benchmarks for fair and consistent evaluation. Future work includes deeper analysis of each objective to understand their failure cases, scaling to larger datasets, more challenging multi-step tasks, and temporal compression methods.


\newpage
\bibliography{iclr2026_conference}
\bibliographystyle{iclr2026_conference}

\newpage
\appendix
\section{Appendix}

\subsection{Hyperparameters}
\label{hyperparameters}

Note that LIV objectives suffered from instability and required a learning rate of $1e-5$ to get meaningful results.

\begin{table}[htbp]
\centering
\caption{Training hyperparameters for the SigLIP-LoRA model.}
\label{tab:training_hyperparams}
\renewcommand{\arraystretch}{1.1}
\begin{tabular}{lcc}
\toprule
\textbf{Parameter} & \textbf{Symbol} & \textbf{Value} \\
\midrule
Total samples & $N_{\text{samples}}$ & $5\times10^{4}$ \\
Validation split & $r_{\text{val}}$ & $0.1$ \\
LoRA rank & $r$ & $16$ \\
LoRA alpha & $\alpha_l$ & $32$ \\
Epochs & $T$ & $5$ \\
Batch size & $B$ & $32$ \\
Learning rate & $\eta$ & $1\times10^{-4}$ \\
Minimum learning rate & $\eta_{\min}$ & $1\times10^{-6}$ \\
Scheduler & -- & Cosine Annealing \\
Margin (triplet) & $\alpha $ & 0.3 \\
Negatives (all methods) & $- $ & 3 \\
\bottomrule
\end{tabular}
\end{table}

\subsection{Example of different views in Metaworld}
\label{annotations}

\begin{figure}[ht]
    \centering
    \begin{subfigure}[b]{0.25\textwidth}
        \centering
        \includegraphics[width=\linewidth]{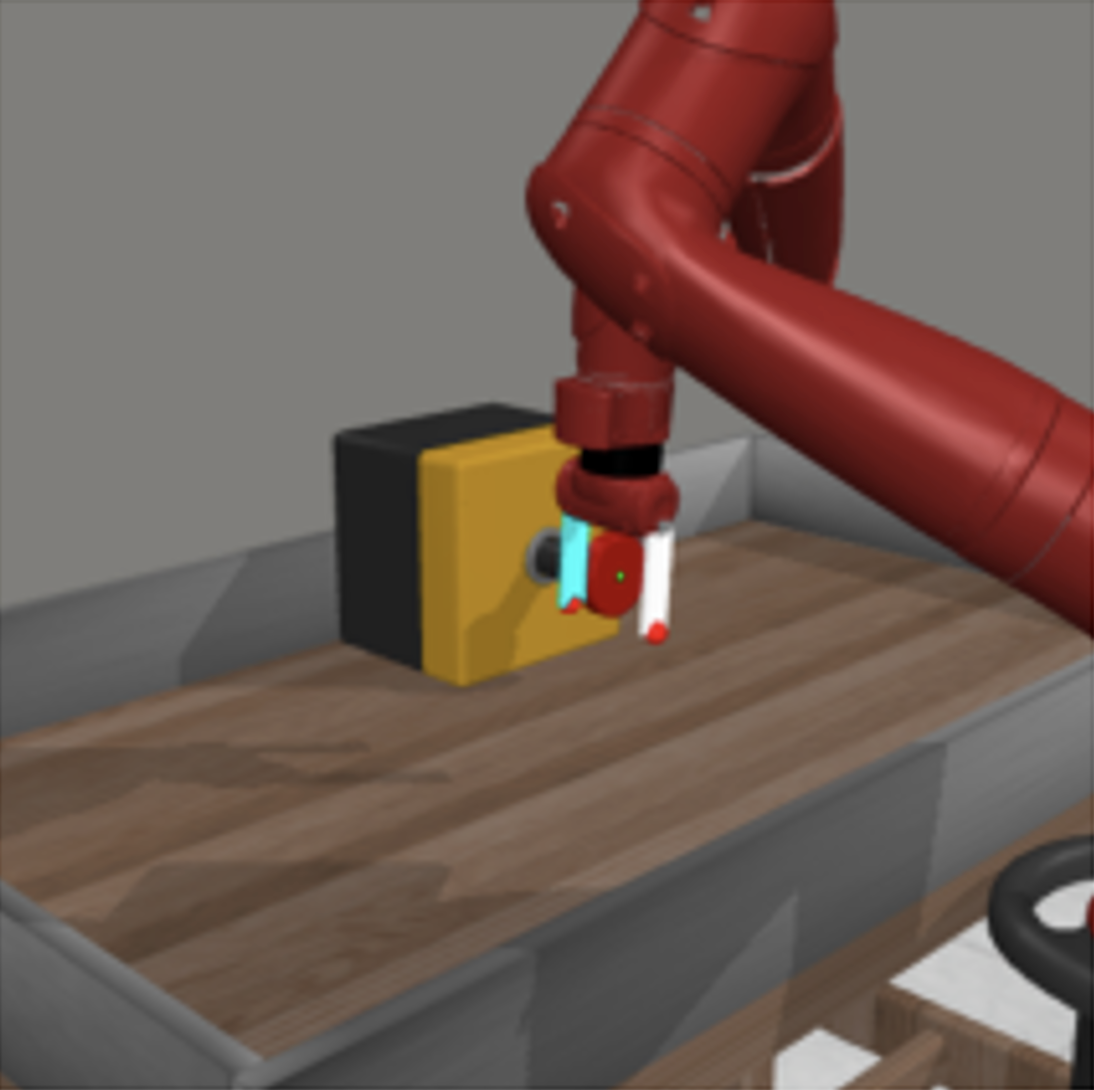} 
        \caption{View 1}
        \label{fig:subfig1}
    \end{subfigure}
    \hfill
    \begin{subfigure}[b]{0.25\textwidth}
        \centering
        \includegraphics[width=\linewidth]{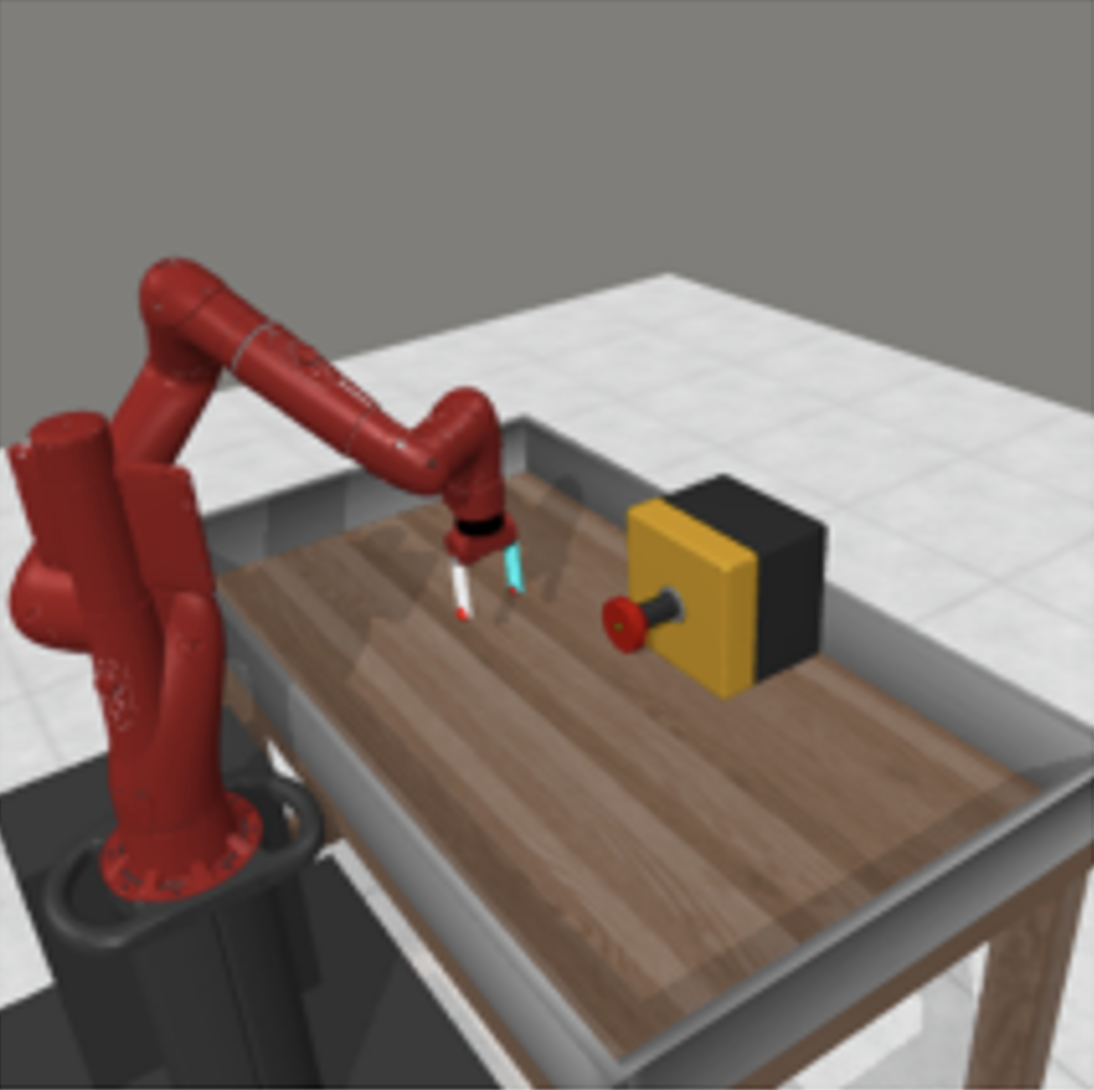}
        \caption{View 2}
        \label{fig:subfig2}
    \end{subfigure}
    \hfill
    \begin{subfigure}[b]{0.25\textwidth}
        \centering
        \includegraphics[width=\linewidth]{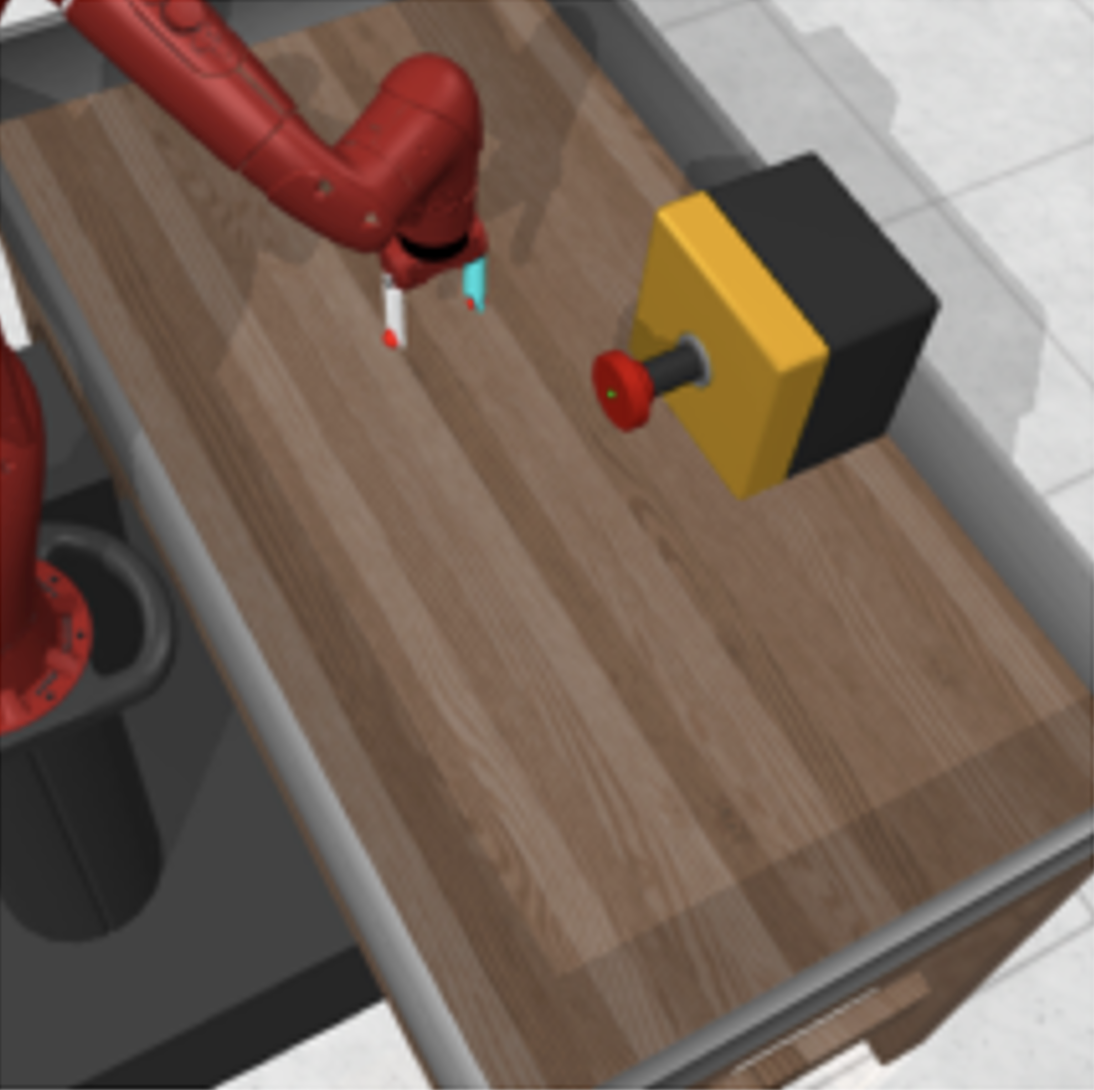}
        \caption{View 3}
        \label{fig:subfig3}
    \end{subfigure}

    \caption{The \texttt{button-press} environment seen from all 3 views. This example illustrates the occlusion problem often encountered in robotics, highlighting the importance of multi-view training and spatial planning.}
    \label{fig:three_subfigs}
\end{figure}

\FloatBarrier
\subsection{Language goal}
\label{language_goal}

\begin{table*}[ht]
\centering
\caption{Language descriptions of the goal for all Metaworld tasks used in our experiments. Underlined tasks are left out the training set, and bold tasks are used for out-of-domain testing.}
\label{tab:task_descriptions}
\begin{small}
\def\arraystretch{0.95}
\begin{tabular}{@{}ll@{}}
\toprule
\textbf{Task} & \textbf{Description} \\
\midrule
assembly-v3 & pick up a nut and place it onto a peg \\
basketball-v3 & dunk the basketball into the basket \\
bin-picking-v3 & grasp the green object from its bin and place it into the other bin \\
box-close-v3 & grasp the cover and close the box with it \\
\underline{button-press-topdown-v3} & press the button on the box \\
\underline{button-press-topdown-wall-v3} & bypass the wall and press the button on the box \\
\underline{\textbf{button-press-v3}} & press the button \\
\underline{button-press-wall-v3} & bypass the wall and press the button \\
coffee-button-v3 & push the button on the coffee machine \\
dial-turn-v3 & rotate the dial 180 degrees \\
disassemble-v3 & pick the nut out of a peg \\
door-close-v3 & close the door with a revolving joint \\
door-lock-v3 & lock the door by rotating the lock clockwise \\
\underline{\textbf{door-open-v3}} & open the door with a revolving joint \\
door-unlock-v3 & unlock the door by rotating the lock counter-clockwise \\
hand-insert-v3 & insert the gripper into the hole \\
faucet-open-v3 & turn on faucet \\
faucet-close-v3 & turn off faucet \\
hammer-v3 & hammer the screw on the wall \\
handle-press-side-v3 & press the handle down \\
handle-press-v3 & press the handle down \\
handle-pull-side-v3 & pull the handle up \\
handle-pull-v3 & pull the handle up \\
lever-pull-v3 & pull the lever down 90 degrees \\
pick-place-wall-v3 & pick the red object, bypass the wall and place the object on the target \\
pick-out-of-hole-v3 & pick up the object from the hole \\
pick-place-v3 & pick and place the red object to the goal \\
plate-slide-v3 & slide the puck into the hockey net \\
plate-slide-side-v3 & slide the puck into the hockey net \\
plate-slide-back-v3 & get the puck from the hockey net \\
plate-slide-back-side-v3 & get the puck from the hockey net \\
peg-insert-side-v3 & insert the peg in the hole \\
peg-unplug-side-v3 & unplug the peg from the hole \\
soccer-v3 & kick the soccer ball into the goal \\
stick-push-v3 & grasp the blue stick and push the box using the stick \\
stick-pull-v3 & grasp the blue stick and pull the box using the stick \\
push-v3 & push the red object to the green objective \\
push-wall-v3 & bypass the wall and push the object behind the wall \\
push-back-v3 & push the red object back to the green objective \\
reach-v3 & reach the red goal position \\
reach-wall-v3 & bypass the wall and reach the red goal position \\
shelf-place-v3 & pick up the blue object and place it onto the shelf \\
sweep-into-v3 & sweep the object into the hole \\
sweep-v3 & sweep the object off the table \\
window-open-v3 & open the window using the handle \\
window-close-v3 & close the window using the handle \\
drawer-close-v3 & close the green drawer \\
\underline{\textbf{drawer-open-v3}} & open the green drawer \\
coffee-push-v3 & push the mug to the red goal under the coffee machine \\
coffee-pull-v3 & pull the mug to the green goal \\
\bottomrule
\end{tabular}
\end{small}
\end{table*}

\FloatBarrier
\subsection{Reward over expert trajectories}
\label{reward_over_trajectories}
\begin{figure}[ht]
    \centering
    \includegraphics[width=0.7\linewidth]{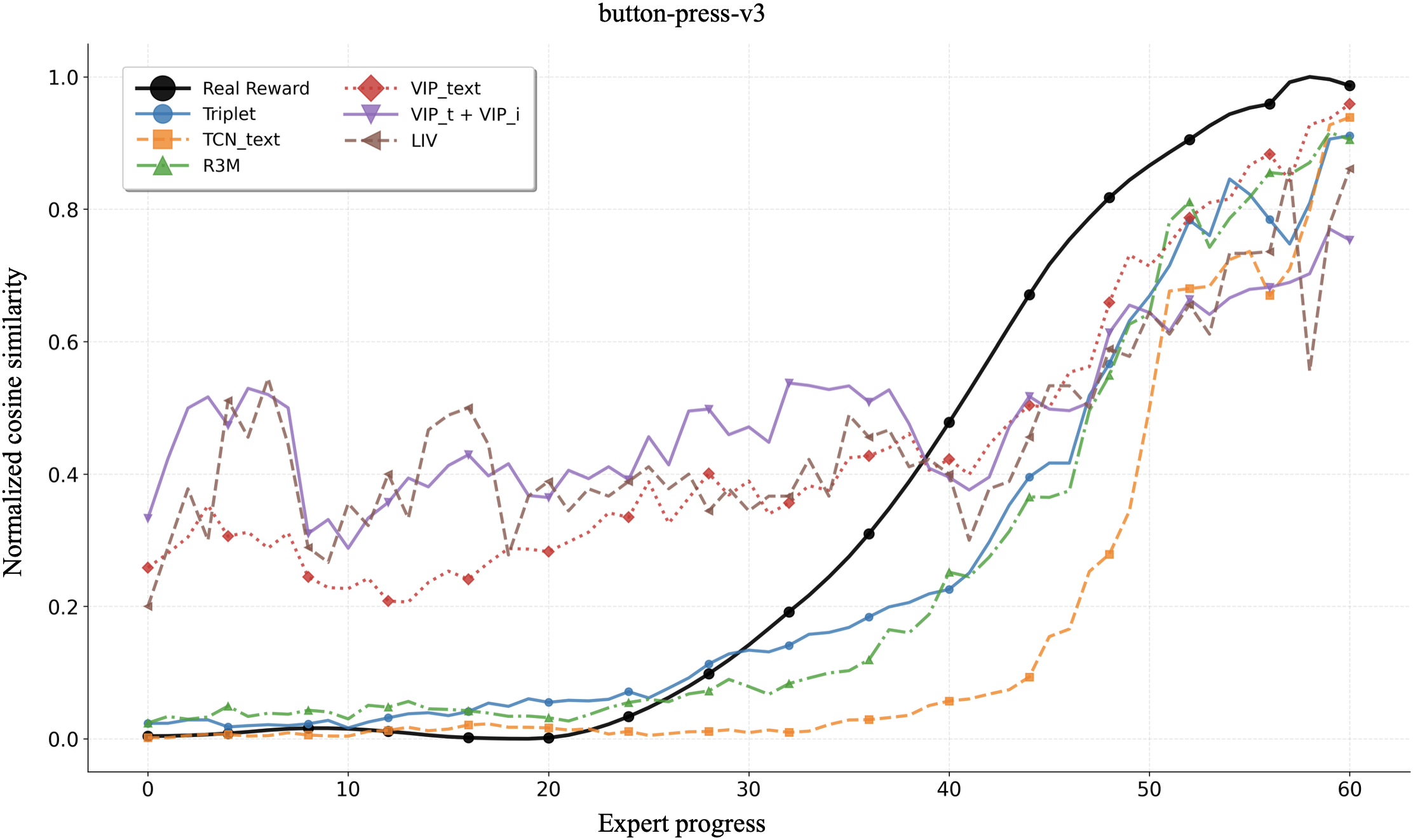}
    \caption{Progression of the reward generated by each model over an expert trajectory in the button-press-v3 environment.}
    \label{fig:reward_button}
\end{figure}

\begin{figure}[ht]
    \centering
    \includegraphics[width=0.7\linewidth]{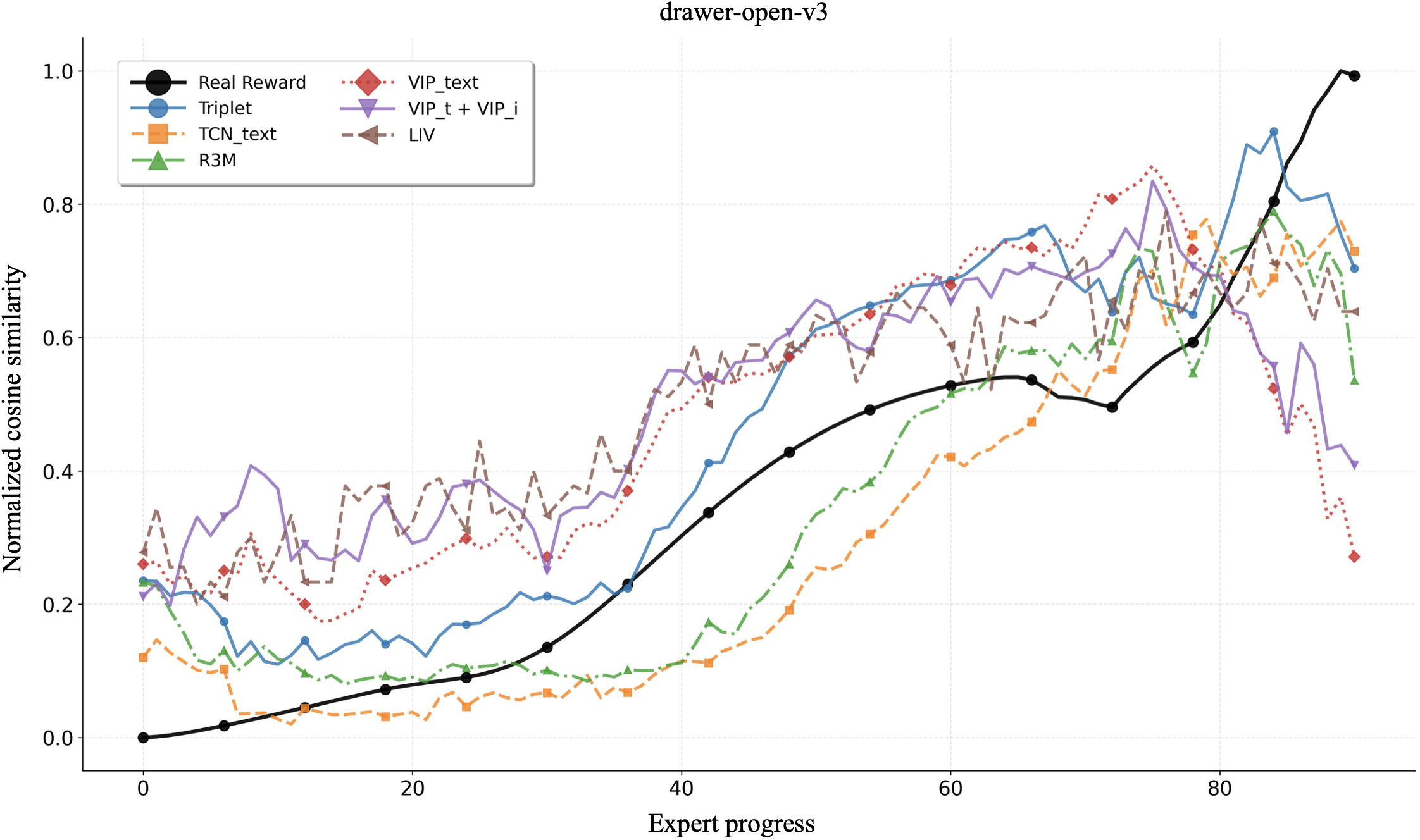}
    \caption{Progression of the reward generated by each model over an expert trajectory in the drawer-open-v3 environment.}
    \label{fig:reward_drawer}
\end{figure}

\begin{figure}[ht]
    \centering
    \includegraphics[width=0.7\linewidth]{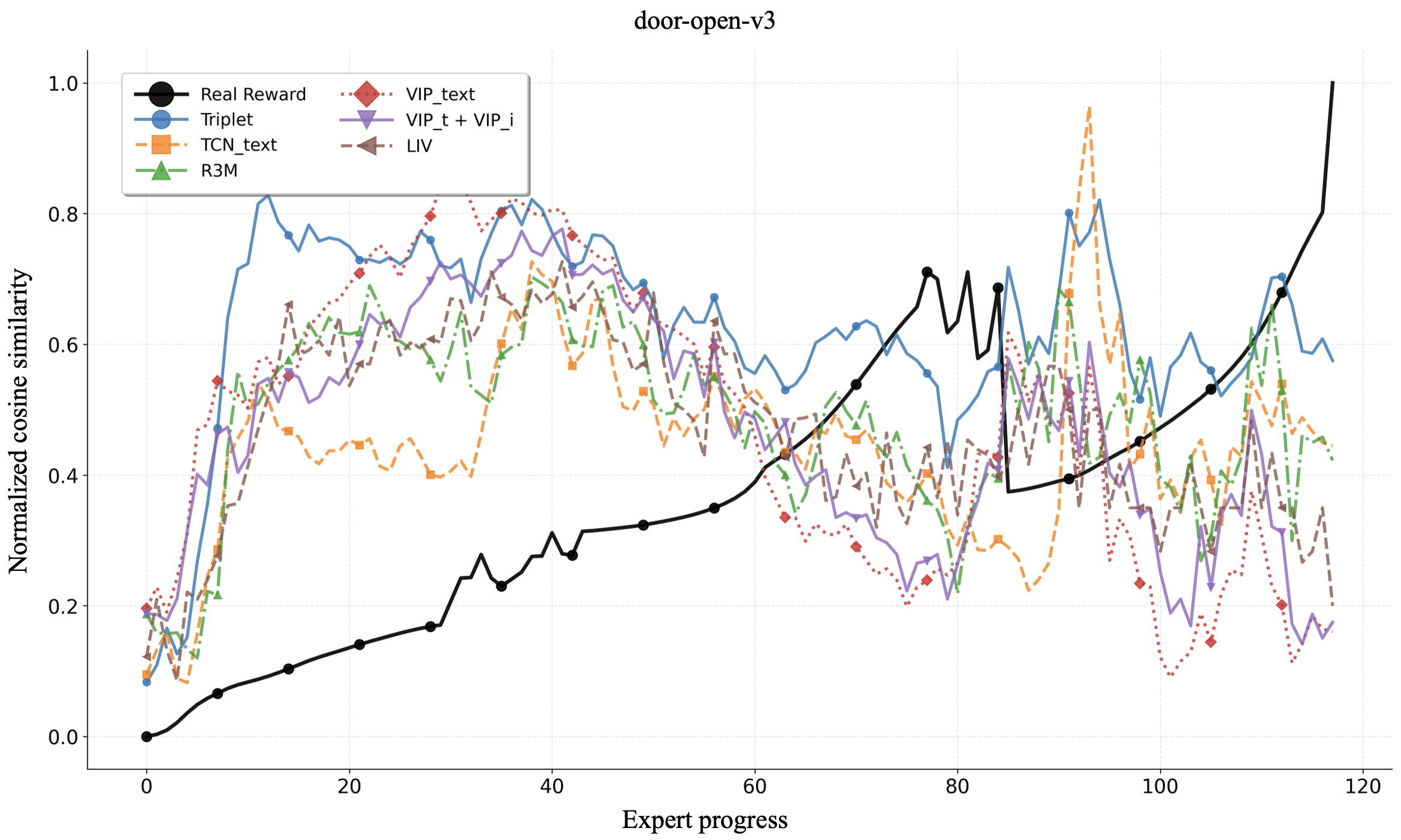}
    \caption{Progression of the reward generated by each model over an expert trajectory in the door-open-v3 environment. }
    \label{fig:reward_door}
\end{figure}

\end{document}